\pdfoutput=1
\documentclass[11pt,table]{article}

\usepackage[]{ACL2023}
\usepackage{times}
\usepackage{latexsym}
\usepackage{hyperref}
\usepackage{microtype}
\usepackage{lipsum} % For generating placeholder text
\usepackage{graphicx}
\usepackage{pifont}% http://ctan.org/pkg/pifont
\usepackage{booktabs}
\usepackage{makecell}
\usepackage{subfigure}
\usepackage{booktabs}
\usepackage{soul,color}

\usepackage[T1]{fontenc}
\usepackage[utf8]{inputenc}

\usepackage{microtype}

\usepackage{inconsolata}

\usepackage{csquotes}

\title{Large Language Model Adaptation for Financial Sentiment Analysis}

\author{Pau Rodriguez Inserte\textsuperscript{1}, Mariam Nakhlé\textsuperscript{1,2}, Raheel Qader\textsuperscript{1\thanks{\vspace{3ex} Corresponding author.}}, Gaëtan Caillaut\textsuperscript{1}, Jingshu Liu\textsuperscript{1} \\
  \textsuperscript{1}Lingua Custodia, Paris, France \\
  \texttt{firstname.lastname@linguacustodia.com} \\
  \textsuperscript{2}Univ. Grenoble Alpes, CNRS, Grenoble INP, LIG, 38000 Grenoble, France 
}

\begin{document}
\maketitle
\begin{abstract}
	Natural language processing (NLP) has recently gained relevance within financial institutions by providing highly valuable insights into companies and markets' financial documents. However, the landscape of the financial domain presents extra challenges for NLP, due to the complexity of the texts and the use of specific terminology. Generalist language models tend to fall short in tasks specifically tailored for finance, even when using large language models (LLMs) with great natural language understanding and generative capabilities. This paper presents a study on LLM adaptation methods targeted at the financial domain and with high emphasis on financial sentiment analysis. To this purpose, two foundation models with less than 1.5B parameters have been adapted using a wide range of strategies. We show that through careful fine-tuning on both financial documents and instructions, these foundation models can be adapted to the target domain. Moreover, we observe that small LLMs have comparable performance to larger scale models, while being more efficient in terms of parameters and data. In addition to the models, we show how to generate artificial instructions through LLMs to augment the number of samples of the instruction dataset.

\end{abstract}

\section{Introduction}\label{introduction}

Natural Language Processing (NLP) has become an increasingly important field in the financial industry, with applications ranging from sentiment analysis and named entity recognition to question answering. Information retrieved using machine learning from financial reports, news or posts in social media can be used as indicators of companies' performance or as insights of a market. Many industry actors are interested in extracting this information to use it as a resource that can provide them with a competitive advantage, such as firms forecasting internal future benefits and losses, investors extracting differential information for trading purposes or any practitioner interested in tracking financial assets.
Nevertheless, some characteristics of financial text make these tasks especially challenging for models that have been trained on general domain data. The use of specific terminology along with the high complexity of the documents, leads these generalist language models to underperform on financial tasks, which suggests that domain adaptation might be required to improve accuracy of interpretation and analysis.

Furthermore, the rapid evolution of large language models (LLMs) and their proven capabilities for NLP tasks has made them stand out and become an interesting option to study. Due to the fact that even the best general language models fall short for some financial tasks, some proposals have been recently presented for a financial domain adaptation of LLMs. These models tailored for finance, such as BloombergGPT~\cite{bloomberggpt}, have been introduced as multitasking generative models specifically designed for financial text understanding and generation. However, these fine-tuned models still show room for improvement, both in performance and in the efficiency of the proposed training strategies.

This paper tackles various aspects of adapting LLMs to the financial domain. In particular, we explore diverse strategies of domain adaptation and fine-tuning of LLMs for financial sentiment analysis, and conduct a series of experiments over two different foundation models. The study focuses particularly on smaller manageable models, up to 1.5B parameters, in order to explore the possibilities of models that can be accessible with relatively low hardware requirements. Although the adapted models are smaller than the current state-of-the-art ones, results show that they achieve similar or higher performance.
In addition, a curated data collection with two main datasets is also presented. One constructed with financial documents and reports, and the other a set of instructions for financial tasks. We show, step by step, the process of creating these datasets and particularly focus on the use of more powerful LLMs to generate synthetic instructions to fine-tune smaller LLMs. Finally, apart from the main focus of the study which is on financial sentiment analysis, other tasks have also been evaluated to analyze the multitasking capabilities of our models.

\section{Related work}\label{related-work}

% \hl{- - SHOULD BE REMOVED - - In this section, the evolution of sentiment analysis in the financial domain is studied, starting from deterministic methods based on dictionaries specifically tailored for finance and concluding with state-of-the-art models and their current weaknesses. Moreover, it includes a definition of large language foundation models and presents the LLMs used during this project.

% Sentiment analysis is one of the most common use cases of NLP. In this task, a model classifies a text according to the sentiment detected, usually between \textit{positive}, \textit{negative} and \textit{neutral}. However, while any sentiment analysis model would be capable of undertaking financial sentiment analysis, an adaptation is required. One example that clearly illustrates why not doing a proper adaptation for finance can be problematic, is the polysemy of the word \textit{good}, which in a general context would be associated to a positive sentiment. However, in finance, the term is frequently used as a synonym of \textit{tangible assets}, which has a neutral connotation. In this subsection, the evolution of financial sentiment analysis approaches before LLMs is presented.}

Sentiment analysis is one of the most common use cases of NLP\@. In this task, a model classifies a text according to the sentiment detected, usually between \textit{positive}, \textit{negative} and \textit{neutral}. However, while any sentiment analysis model would be capable of undertaking financial sentiment analysis, an adaptation is required. In this section, the evolution of sentiment analysis in the financial domain is studied using models based on Transformers~\cite{transformers}.

\textbf{FinBERT}\footnote{Several models under the name of FinBERT exist, however in this work we only discuss the first model.}~\cite{finbert} is based on the idea of training a BERT~\cite{bert} model in two steps to adapt it to the financial domain and the sentiment analysis task. The first step consists of further pre-training the model on financial documents, as this strategy has already be proven to be effective~\cite{ulmfit} for domain adaptation. This step aims at helping the model to understand financial terminologies better than the base model. Authors used a subset of Reuters' TRC24 dataset\footnote{\url{https://trec.nist.gov/data/reuters/reuters.html}}, a collection of news articles published by Reuters that was filtered with keywords related to finance, to fine-tune the model. In the second step, the model is prepared for the sentiment analysis task by adding a dense layer to the last hidden state of the classification token \verb|CLS| of the encoder-based architecture, a recommended practice for classification with BERT\@. This task is fine-tuned using the Financial PhraseBank (FPB)~\cite{financial-phrasebank}, a financial sentiment analysis dataset. FinBERT presents remarkable results on financial sentiment analysis, outperforming the state-of-the-art. Nevertheless, the model is strongly limited to sentiment analysis and underperforms greatly on other tasks.
% \hl{Similar models, also called FinBERT despite being from different authors, were developed following a similar strategy. FinBERT} \cite{finbert-yang}, \hl{also continued training a BERT base model on a financial dataset, constructed with different documents than Acraci's, and fine-tuned it on different sentiment analysis tasks: FPB, but also FIQA and AnalystTone. A third FinBERT} \cite{finbert-liu} \hl{proves that BERT-based models can also be fine-tuned on other tasks if the final layers are fine-tuned for them.}

\subsection{Base large language models}
Recent advances in the field of large language models (LLMs) have shown that these models can achieve remarkable capabilities in understanding complex natural language. They are also capable of performing zero-shot and few-shot learning, in which they can generate accurate responses for tasks that they have not seen during training~\cite{in-context-learning}. This makes LLMs a great choice in multitask settings where one model is expected to perform several tasks. Most of today's LLMs are based on Transformer models~\cite{transformers}, typically set in decoder-only architectures.

% Significant progress in the field of large language models (LLMs) has demonstrated their remarkable capabilities across various NLP tasks. These models have proven great performance in zero-shot and few-shot learning settings, in which they are able to generate accurate outputs with minimal examples, even when the models have not been exposed to these specific examples during their training~\cite{in-context-learning}. Consequently, these models have the ability to multitask across NLP tasks through text generation. LLMs are based in Transformer models~\cite{transformers}, and while FinBERT and BERT had an encoder-only architecture, these models are commonly structured with only decoders.

Training of LLMs is typically split in two stages. The first part of the training is the most computationally expensive since the model is trained using large amounts of text. For this reason, conducting the training of a LLM from scratch requires high computational resources. Nevertheless, many research groups and companies are releasing these models to the public to be used as base or foundation for other models, to enable research to move forward. Using these pre-trained models is highly beneficial for researchers with fewer data or hardware resources, as they can be used as a starting point for fine-tuning on specific tasks, such as chatting, following instructions or giving outputs in a specific style or format. Although most LLMs are trained on general domain data, there have been a few works recently to adapt LLMs to the financial domain. In the next subsections two such work are reviewed.
% The choice of the foundation models used for this project was based on their performance on general tasks, openness, conditions of their license and size of the model, a constraint set by the GPU used for the experiments.

\subsection{Financial large language models}
\textbf{BloombergGPT.} One of the first decoder-only LLMs trained specifically for finance is BloombergGPT~\cite{bloomberggpt}, a model of 50B parameters based on BLOOM's architecture~\cite{bloom-short}. The corpus collected for the training of this LLM consisted in the combination of 363 billion tokens from financial documents with 345 billion tokens from general purpose datasets. The model was trained from scratch, without using any foundation model as a base, with the objective of predicting the next token of the documents, and without fine-tuning on instructions. However, the results presented by BloombergGPT are far from the ones achieved by other models, some of them of a much smaller scale. In addition, the results reported did not outperform other generalist LLMs, as we will show later in this paper.

\textbf{FinMA.} The open model FinMA from PIXIU's framework~\cite{finma} introduced by ChanceFocus reported better scores on several financial tasks than larger generalist LLMs, such as GPT-4~\cite{gpt4}, and BloombergGPT\@. They used LLaMA~\cite{llama} as the pre-trained model and fine-tuned it with instructions tailored for financial multitasking. The instruction dataset consists of texts formed by an instruction, an input and an answer. The dataset includes a data augmentation strategy in which the inputs of those tasks with few  samples were used with 10 different instructions. This augmentation strategy, while increasing the number of samples in the dataset, did not increase its diversity as the same set of 10 instructions were always repeated.

In the same paper in which FinMA was presented, the PIXIU framework also included FLARE, a financial evaluation benchmark. This benchmark has been used to evaluate the experiments carried out in this project.

\subsection{Financial benchmark}
For the evaluation of large language models, the FLARE benchmark\footnote{\url{https://github.com/chancefocus/PIXIU}} from PIXIU framework has been used. The tasks of this benchmark which are relevant to our work are presented below.

\textbf{Financial Sentiment Analysis.} Financial sentiment analysis task over two different benchmarks, the Financial Phrase Bank (FPB)~\cite{financial-phrasebank} and FIQA-SA~\cite{fiqasa}.

\textbf{News Headline Classification.} Headlines task contains 9 different subtasks, each one associated with 9 different gold questions, in which the expected answers are \enquote{yes} or \enquote{no}. The inputs analyzed are gold news from the Gold dataset~\cite{gold}.

\textbf{Named Entity Recognition.} NER task is based on detecting financial named-entities in U.S. public agreements in the~\cite{ner} dataset. The tagged entities correspond to people, organizations and locations.

\section{Methodology}\label{methodology}

In this section, we describe the methods designed to conduct the experiments.
First, we list the foundation models that are used as a part of this project. Then, two new dataset collections are introduced, one with data based on documents and the second with instructions. We also give details of designing a data augmentation strategy for the instructions as well as the description of the training  process carried out to fine-tune the foundation models.

\subsection{Foundation models}
As stated earlier, the focus in this work is on smaller sized models that can be adapted to achieve performance of larger models. The two models that we use are listed below:

\textbf{OPT.} Meta AI’s large language models suite OPT (Open Pre-trained Transformers)~\cite{opt-llm} were presented as a collection of 9 models ranging from 125M to 175B parameters, being one of the first publicly available LLMs.

\textbf{Pythia}. EleutherAI presented Pythia~\cite{pythia-llm}, a suite of decoder-only language models with sizes ranging from 70M to 12B parameters. These models are trained on the Pile dataset~\cite{the-pile}, a curated collection of English texts from a wide variety of sources.

\subsection{Datasets}
In order to train LLMs, two main different approaches can be taken with respect to data. When a model is trained from scratch, the data used are collections of documents, for which the model has the objective of predicting the next token. This is usually the training carried out to obtain foundation LLMs. However, these models can be further pre-trained for domain adaptation, in the same way that FinBERT was trained. This approach is based on the idea of continuing the training of the model with financial documents to shift from a general to a financial language model.
Moreover, it has been proven that large language models can improve their performances, especially on unseen (or zero-shot) tasks by fine-tuning them to follow instructions. For this fine-tuning method, the training objective is the same, predicting the next token of the text, with the only difference being that the format of this data relies on an \enquote{instruction}, \enquote{input}, \enquote{answer} format.
For this project, one dataset was collected for each of these two training strategies. In addition, the instruction-based dataset was augmented artificially with samples generated from another LLM (LLaMA 2 13B~\cite{llama2-llm}).

\textbf{Document dataset.} The collection of documents used to further pre-train the base LLMs is a combination of general and financial documents from different sources. The purpose of this mixture is to add diversity to the training data, with finance being the most represented domain. Having general data in the training set prevents the model to completely drift the domain and result in a model that is unable to understand general language. The data sources of these documents are described below:
\begin{itemize}
	\item \textbf{EDGAR Files} (Financial). EDGAR is the Electronic Data Gathering, Analysis, and Retrieval system online platform operated by the SEC\footnote{\url{https://www.sec.gov/edgar/searchedgar/companysearch}} (United States Securities and Exchange Commission). It is used by companies to electronically file registration statements, periodic reports, and other forms required by the SEC\@. The database of these documents is open to everyone, allowing the retrieval of high-quality financial text.
	\item \textbf{Reuters News} (Financial). Reuters is a news agency specializing in business and finance that released Reuters Corpora, a collection of financial news made available for use in NLP research. The collection used in this dataset is TRC2 (Thomson Reuters Text Research Collection), that contains more than 1.8 million news.
	\item \textbf{In-house Dataset} (Financial). As a part of this project, a diverse collection of in-house financial text has been obtained. The text in this dataset is mostly at sentence level, as they were originally used for machine translation. This is the only private set used for the project.
	\item \textbf{The Pile} (General). The Pile~\cite{the-pile}, from EleutherAI, is a dataset that comprises 22 diverse high-quality subsets, several of which originate from academic or professional sources. The idea behind this dataset's construction is that diversity enhances general cross-domain knowledge and downstream generalization capability of large language models. It includes data from general news to scientific articles, code, etc\ldots The proportions of these subsets are kept as-is in the sub-sample used for this project.
\end{itemize}

The lengths of the documents of this dataset had to be adapted to the models’ context length, which corresponds to the longest sequence of tokens that the model can support. In this project the context was limited to 2048 tokens. The pre-processing of the dataset consisted in concatenation of all documents from the same source, using a special token (\verb|<endoftext>|) to separate them. The long concatenated text is then sliced in blocks of 2048 tokens, which are  mixed and shuffled with all the other blocks of the dataset. Since some datasets used in this project are extremely large, we decided to take a smaller proportion from each one. The summary of the ratio used for each partition is shown in Table~\ref{tab:docs-data}.

\begin{table}[ht]
	\centering
	%\begin{tabular}{lccc}
	%	\toprule
	%	\textbf{Subset} & \textbf{Domain} & \textbf{Num. tokens} & \textbf{Prc. (\%)} \\
	%	\midrule
	%	EDGAR Files   & Financial & 100,000              & 25.7               \\
	%	Reuters News    & Financial & 36,000               & 9.3                \\
	%	In-house        & Financial & 38,000               & 9.7                \\
	%	The Pile     & General & 215,000              & 55.3               \\
	%	\midrule
	%	\textbf{Total}  & & 389,000              & 100                \\
	%	\bottomrule
	%\end{tabular}
 	\begin{tabular}{llrr}
		\toprule
		\textbf{Subset} & \textbf{Domain} & \textbf{\# Tokens} & \textbf{\%} \\
		\midrule
		EDGAR   & Finance & 100k              & 25.7               \\
		Reuters    & Finance & 36k               & 9.3                \\
		In-house        & Finance & 38k               & 9.7                \\
		The Pile     & General & 215k              & 55.3               \\
		\midrule
		\textbf{Total}  & & 389k              & 100                \\
		\bottomrule
	\end{tabular}
	\caption{Proportion and absolute number of tokens taken from each dataset.}\label{tab:docs-data}
\end{table}

\textbf{Instruction-based dataset.}
Instruction fine-tuning is a strategy used to improve LLMs' performance for specific tasks by teaching them to follow specific format of questions and answers. LLMs learn by being trained on this specific format of text, while keeping the same training objective, predicting the next sequence of tokens. Fine-tuning on instructions is the most common technique to adapt foundation models to specific use-cases, mainly because this method not only improves performance on the trained tasks, but also augments zero-shot and few-shot capabilities. Models trained on instructions are usually consistent in the format in which data is presented. In Table~\ref{tab:template-instr} the format used for our dataset is displayed.

\begin{table}[ht]
	\small
	\centering
	\begin{tabular}{l}
		\toprule
		\textbf{Template}                                                            \\
		\midrule
		\texttt{\#\#\# Instruction: Description of the task} \\
		\texttt{\#\#\# Input: Input to analyze}                                    \\
		\texttt{\#\#\# Answer: Answer to predict}                                  \\
		\bottomrule
	\end{tabular}
	\caption{Template for instructions.}\label{tab:template-instr}
\end{table}

To create the instruction dataset, we used the instructions dataset published by PIXIU that targeted the FLARE benchmark. However, this dataset  has poorly curated prompts and includes a suboptimal data augmentation strategy. For instance, certain parts of the dataset have been up-sampled by using ten different instructions over the same input. Despite having more samples, the up-sampled version of dataset lacks diversity which may lead to poor performance as discussed by~\citet{lima}. For this reason, the dataset proposed in this project has been designed from scratch, only reusing the unique \textit{input - answer} pairs from a down-sampled version of PIXIU's dataset that includes a single instruction for each input.
\iffalse
	\begin{table}[ht]
		\centering
		\begin{tabular}{lcc}
			\toprule
			\textbf{Subset} & \textbf{\hl{PIXIU}} & \textbf{\hl{Down-sampled}} \\
			\midrule
			FPB             & 48,380              & 4,838                      \\
			FiQA-SA         & 11,730              & 1,173                      \\
			NER             & 6,090               & 609                        \\
			Headline        & 102,708             & 102,708                    \\
			FinQA           & 8,242               & 8,242                      \\
			ConvFinQA       & 3,892               & 3,892                      \\
			BigData22       & 7,164               & 0                          \\
			ACL18           & 27,053              & 0                          \\
			CIKM18          & 4,967               & 0                          \\
			\midrule
			\textbf{TOTAL}  & \textbf{220,226}    & \textbf{121,462}           \\
			\bottomrule
		\end{tabular}
		\caption{Comparison between PIXIU's dataset and the one proposed. The dataset is down-sampled and the prompts are modified.}\label{tab:instr-data}
	\end{table}
\fi

\begin{table}[]
    \begin{tabular}{lccc}
		\hline
		\textbf{Subset} & \textbf{Instr}   & \textbf{PIXIU}   & \textbf{Augm}    \\ \hline
		FPB                              & 4,838                             & 48,380                            & 6,633                             \\
		FiQA-SA                          & 1,173                             & 11,730                            & 2,825                             \\
		NER                              & 609                               & 6,090                             & 2,609                             \\
		Headline                         & 102,708                           & 102,708                           & 102,708                           \\
		FinQA                            & 8,242                             & 8,242                             & 8,242                             \\
		ConvFinQA                        & 3,892                             & 3,892                             & 3,892                             \\
		BigData22                        & 0                                 & 7,164                             & 0                                 \\
		ACL18                            & 0                                 & 27,053                            & 0                                 \\
		CIKM18                           & 0                                 & 4,967                             & 0                                 \\ \hline
		\textbf{TOTAL}  & \textbf{121,462} & \textbf{220,226} & \textbf{126,909} \\ \hline
    \end{tabular}
    \caption{Comparison of the instructions used for each task in the original instructions dataset (\textit{Instr}) with one input-answer by instruction, the up-sampled version (\textit{PIXIU}), and the dataset augmented by LLM inference (\textit{Augm}).}\label{tab:instr-data}
\end{table}

\textbf{Instruction data augmentation.}
%In order to replace the data augmentation strategy proposed by PIXIU, a method based on using the generative capabilities of LLMs has been designed. The goal of this method was to add completely new instructions to the dataset for those tasks that are less represented. The main idea behind these synthetic instructions is to bring a new input to the dataset so the model has more diverse examples during training.
The main idea behind instruction data augmentation is to bring new inputs to the dataset, so the model has more diverse examples to learn from.
Two different methods have been defined for generating these instructions dependent on the target task. For sentiment analysis, the model has to generate an input for a given sentiment given an example with that label. This strategy has been used to augment both FPB and FIQA-SA subsets. In Table~\ref{tab:template-sa} of Appendix~\ref{sec:inst_data_aug} the template used for this task is presented. %the template used is presented, where $\{ xi, yi \}$ is an \textit{input-answer} pair sampled from one of the two subsets.

% The design of the prompt is based on the following ideas:
% \begin{itemize}
%     \item By adding the sentiment in which the new sentence has to be classified in the prompt, the language model does not need to carry out this classification task. By this, it is avoided a possible mismatch between the new \textit{input-answer} pair that would propagate an error on the fine-tuned model. Moreover, adding the class label at the beginning of the instruction has positive effects as shown in the self-instruct method~\cite{self-instruct}.
%     \item Three different elements lead the generative model to follow the style of the dataset that is being augmented: the example given from the targeted subset; the instruction to reuse elements from the example, which allows the model to make use of names of companies, numbers and other financial terms; the explicit instruction to generate a sentence with the given financial sentiment.
%     \item By asking the models to generate the sentence in a specific format, the input is easier to retrieve afterwards. This is important because language models tend to add extra text to their output if they have not been fine-tuned for the task. This effect is especially visible in models fine-tuned for chat.
%     \item The example given is randomly sampled from the training set with the purpose of preserving the original distribution on the answers generated.
% \end{itemize}

For NER, since it is not a sequence classification task, the inference method is different. The first solution proposed was based on letting the model generates both the new sentence and its NER tags at the same time, only guiding the model by including a few examples in the prompt. However, the variety of the sentences generated by the model was too short and the tags were incorrect, indicating that the task was too hard for the model. Our solution was to use existing unlabeled sentences, which reduced the generative task to a tagging process. The sentences used for this augmentation were in-house financial sentences. Moreover, in this case the example given to the model is fixed in order to make sure all types of entities are present in the prompt. The format of the tagging was chosen using prompt engineering. In Table~\ref{tab:template-ner} of Appendix~\ref{sec:inst_data_aug}, the template used for NER data augmentation is shown. The Headline task was not augmented since it had enough samples, even when considering that there are 9 subtasks in the benchmark.

Using the above-mentioned two templates, new inputs are inferred to be added to the instructions dataset. The model used for the generation of these new samples is LLaMA-2-13B, quantized in 4-bits to reduce the GPU memory required. Table~\ref{tab:instr-data} shows a comparison of the number of samples targeting each task before and after the augmentation. The decision on the number of synthetic samples generated was taken considering the number of original samples. The reason behind not generating even a larger number of instructions is that despite their high-quality, artificial samples could introduce some noise to the dataset by generating sentences too different from the original distribution, introducing erratic \textit{input-answer} pairs or NER tags or to duplicate some inputs after several iterations.
In Table~\ref{tab:instr-data} there is a comparison between PIXIU's dataset before and after down-sampling as well as after the data augmentation.

\iffalse
	\begin{table}[ht]
		\centering
		\begin{tabular}{lcc}
			\toprule
			\textbf{Task}  & \textbf{Before augment.} & \textbf{After augment.} \\
			\midrule
			FPB            & 4,838                    & 6,633                   \\
			FiQA-SA        & 1,173                    & 2,825                   \\
			NER            & 609                      & 2,609                   \\
			Headline       & 102,708                  & 102,708                 \\
			FinQA          & 8,242                    & 8,242                   \\
			ConvFinQA      & 3,892                    & 3,892                   \\
			\midrule
			\textbf{TOTAL} & \textbf{121,462}         & \textbf{126,909}        \\
			\bottomrule
		\end{tabular}
		\caption{Instructions dataset description before and after data augmentation by LLM inference. Headlines and FiQA are not augmented since they are a big part of the set.}\label{tab:augm-data}
	\end{table}
\fi

\subsection{Training method}
As stated earlier, in this work, we use two pre-trained foundation models, namely Pythia-1.4B and OPT-1.3B, and fine-tune them in two stages as detailed below:

% The fine-tuning of foundation models has been carried out following two different strategies which mainly differ from each other on the dataset used for the training. In further experiments, both strategies are combined. The foundation models fine-tuned for these tasks are Pythia-1.4B and OPT-1.3B.
\begin{itemize}
	\item \textbf{Further pre-training.} The models are fine-tuned to predict the next token of the text in the document-based dataset, following the same idea as in FinBERT and without being fine-tuned on a specific task. The idea here is to tilt the models to become more familiar with the financial domain. Both models are trained on a total of $389,000$ tokens introduced in context blocks of $2,048$ tokens. The models are trained for two epochs, saving 4 checkpoints at every epoch. The best checkpoint is selected for each model.
	\item \textbf{Instruction fine-tuning.} The model is instructed to perform financial tasks using the instructions dataset. Since the length of these instructions is generally shorter than on the document-based dataset, the context length is reduced to $1,000$ tokens to speed up the training. Sequences shorter than this length are padded with a padding token. Instructions longer than $1,000$ tokens are cut off. For instruction fine-tuning, models are trained for 1 epoch.
\end{itemize}
For both set-ups, training is performed with AdamW optimizer~\cite{adamw}, a batch size of 32, and applying gradient accumulation of 4 for training efficiency. The initial learning rate is set to 1e-4, while the weight decay is adjusted to 0.1. These values remained the same for all the conducted experiments. The training of these models was carried out on a H100 GPU.

\section{Results}\label{results}
\subsection{Classical algorithms versus LLMs for financial sentiment analysis}
Prior to conducting an evaluation of our fine-tuned LLMs for financial sentiment analysis, we study the performance of the current state-of-the-art models and classical machine learning algorithms. Compared to LLMs, classical algorithms do not require a lot of computation, they could be easily trained and tested. For the sake of simplicity, the evaluation has been carried out only on the FPB\@. %This dataset has several splits according to the annotators' agreement on the sentiment, but for this experiment they have been mixed and split again randomly in train, validation and test sets. These partitions are proposed by FinBERT, and they are used here in the same way because, since the model has been fine-tuned on this dataset, it is the only way to avoid data leakage. Therefore, the machine learning models are trained with the same FPB training set as FinBERT and all strategies are evaluated on the same test set.
Based on the results in Table~\ref{tab:sota-performance}, the lowest score, unsurprisingly, is obtained by the lexicon approach. Classical machine learning algorithms on the other hand are able to obtain results considerably higher than lexicon, and even match or pass LLM scores in some cases.
% FinBERT outperforms both machine learning and lexicon approaches. This model set the state-of-the-art of financial sentiment analysis dedicated models and was used as reference for more recent LLMs performance. The performance of FinBERT can also be compared to the recent LLMs by using the scores of LLMs reported in PIXIU's paper and BloombergGPT.
Overall conclusions that can be depicted from the results can be summarized as follows:
\begin{itemize}
	\item The domain adaptation and training of FinBERT on this specific task, gives the model an advantage over general models. Comparing FinMA-30B with GPT-4, it can be seen that a smaller model fine-tuned for finance has better performance than a generalist one.
	\item BloombergGPT was a good starting point for financial LLMs. However, its performance on tasks like sentiment analysis is poor. One likely reason is that this model has not been fine-tuned on instructions.
	\item FinMA-30B proves the relevance of fine-tuning on instructions to improve performance on financial tasks. Nevertheless, as mentioned before, the train dataset might be not sufficiently diverse, which may impact the model’s capability in real-world scenario.
\end{itemize}

\begin{table}[ht]
        \small
	\centering
	\begin{tabular}{lcc}
		\toprule
		\textbf{Algorithm and features} & \textbf{Accuracy}  \\
		\midrule
		\multicolumn{2}{l}{\textbf{Lexicon approach}}       \\
		Loughran-McDonald dictionary    & 0.59      \\
		\midrule
		\multicolumn{2}{l}{\textbf{Classical ML algorithms}}          \\
		SVM                             & 0.77           \\
		Naive Bayes                     & 0.73           \\
		XGB                             & 0.80          \\
		\midrule
		\multicolumn{2}{l}{\textbf{Transformers approach}}  \\
		FinBERT                         & 0.85           \\
		GPT-4                           & 0.71              \\
		BloombergGPT                    & -             \\
		FinMA-30B                       & \textbf{0.87}     \\
		\bottomrule
	\end{tabular}
	%\caption{Performance of financial sentiment analysis strategies before LLMs.}\label{tab:sota-performance}
    \caption{Performance of financial sentiment analysis. A comparison between traditional approaches and modern transformer based models.}\label{tab:sota-performance}
\end{table}

% \hl{Added column with F1-scores. TODO: Fill it in and explain (again) why BloombergGPT doesn't have accuracy reported.}
% \begin{table}[ht]
%         \small
% 	\centering
% 	\begin{tabular}{lcc}
% 		\toprule
% 		\textbf{Algorithm and features} & \textbf{Accuracy} & \textbf{F1-score} \\
% 		\midrule
% 		\multicolumn{2}{l}{\textbf{Lexicon approach}}       \\
% 		Loughran-McDonald dictionary    & 0.59 & ?         \\
% 		\midrule
% 		\multicolumn{2}{l}{\textbf{Classical ML algorithms}}          \\
% 		SVM                             & 0.77 & ?            \\
% 		Naive Bayes                     & 0.73 & ?            \\
% 		XGB                             & 0.80 & ?            \\
% 		\midrule
% 		\multicolumn{2}{l}{\textbf{Transformers approach}}  \\
% 		FinBERT                         & 0.85 & ?             \\
% 		GPT-4                           & 0.71 & 0.78              \\
% 		BloombergGPT                    & - & 0.51              \\
% 		FinMA-30B                       & \textbf{0.87} & \textbf{0.88}     \\
% 		\bottomrule
% 	\end{tabular}
% 	%\caption{Performance of financial sentiment analysis strategies before LLMs.}\label{tab:sota-performance}
%     \caption{Performance of financial sentiment analysis. A comparison between traditional approaches and modern transformer based models.}\label{tab:sota-performance}
% \end{table}

\subsection{Financial domain adaptation}

\begin{table*}[ht!]
	\small
	\centering
	\begin{tabular}{p{2.2cm}ccccc}
		\toprule
		                       & \multicolumn{5}{c}{\textbf{F1 scores}}                                                                        \\
		\cmidrule(lr){3-6}
		                & Fine-tuning data                   & FPB         & FIQA-SA       & Headlines       & NER     \\
		\midrule
		\makecell{Pythia-1.4B} & \textit{(base)}              & 0.20          & 0.29         & 0.16          & 0       \\
		                       & \textit{Docs}                 & 0.41          & 0.16         & 0.57          & 0.30    \\
		                       & \textit{Instr}                & 0.82          & 0.73          & 0.93          & 0.59   \\
		                       & \textit{Augmented instr}      & 0.82          & 0.79         & 0.90          & 0.56    \\
		% \arrayrulecolor{black!30}\midrule
		% & \textit{Docs + Instr} & 0.8106 & 0.7721 & 0.9602 & 0.6537 \\
		                       & \textit{Docs + Augmented instr}    & \textbf{0.84}     & \textbf{0.83} & \textbf{0.97} & \textbf{0.69} \\
		\arrayrulecolor{black!}\midrule
		\makecell{OPT-1.3B}    & \textit{(base)}               & 0.19          & 0.48         & 0.29         & 0  \\
		                       & \textit{Docs}                  & 0.13          & 0.58         & 0.39         & 0   \\
		                       & \textit{Instr}                 & 0.84          & 0.77         & 0.93         & \textbf{0.53} \\
		                       & \textit{Augmented instr}       & \textbf{0.86} & 0.79         & \textbf{0.97}  & 0.29 \\
		% \arrayrulecolor{black!30}\midrule
		% & \textit{Docs + Instr} & 0.8011 & 0.7736 & 0.9461 & 0.498 \\
		                       & \textit{Docs + Augmented instr}        & \textbf{0.86} & \textbf{0.81}   & 0.96     & 0.34  \\
		\arrayrulecolor{black!}\bottomrule
	\end{tabular}
	\caption{Comparison of Pythia-1.4B and OPT-1.3B fine-tuned with different strategies. The results reported correspond to the base models without fine-tuning  (\textit{base}), models with document further pre-training (\textit{Docs}), models fine-tuned on instructions (\textit{Instr}), models fine-tuned on augmented instructions dataset (\textit{Augmented instr}), fine-tuning first with documents and then with augmented instructions (\textit{Docs + Augmented instr}).}\label{tab:exp-results}
\end{table*}

In this section we show the impacts of the two stage fine-tuning as well as improvements brought by the artificially augmented instruction dataset. Models are evaluated using a subset of the tasks proposed in the FLARE benchmark: FPB, FIQA-SA, Headlines and NER\@. For the classification tasks (FPB, FIQA-SA and Headlines), the predictions are obtained by forcing the model to generate one of the expected class label. For example, in FPB, this means to choose the next token only amongst the ones needed to generate the labels (\textit{positive}, \textit{negative} or \textit{neutral}), and sticking with the most probable ones (the highest logits).
%the results are obtained by reading the logits at the output of every possible answer. For example, in FPB, this means to choose the maximum value of the logits associated with the tokens \textit{positive}, \textit{negative} and \textit{neutral}.
When evaluating on NER, the generation is not constrained.

% \textbf{Training on documents vs instructions.}
The first experiment that we carry out is to see the effects of fine-tuning on documents versus instructions. Based on the results of Table~\ref{tab:exp-results}, it is clear that performance of both Pythia and OPT models show similar behaviors and that fine-tuning brings significant improvements over the base models. Particularly, instruction fine-tuning improvement is much higher than just further pre-training on documents. This conclusion seems to be aligned with what we observe in the literature of instruction tuning of other domains.

% \textbf{Augmented instructions vs non-augmented instructions datasets.} 
Next, in order to evaluate the effect of augmenting the number of instructions using the strategy designed for this dataset, the models are compared after fine-tuning with the base instructions dataset and with the augmented instructions dataset. The results of using data augmentation are not straightforward. In Table~\ref{tab:exp-results}, it can be seen that the performance of the models augmented instructions is improved for the sentiment analysis tasks, but the scores goes down for the other two. In the case of Headlines, this effect can be caused by the fact that this task is the most represented in the dataset and, by introducing new samples, the model is less focused on this task. For NER, the issue can be explained by the difference between the text of the synthetic samples and the original test set. As explained in previous sections, NER is augmented using in-house data, and even though the chosen sentences were also in the financial domain, the sources are different and that might have introduced errors in the predictions.

Finally, we can test the implications of instruction fine-tuning after further pre-training the model with the financial documents. This simply means that the model is fine-tuned two times. Since the augmented instructions proved to be better than the original instructions, this experiment is conducted on the earlier instruction dataset. As shown in the last row of Table~\ref{tab:exp-results}, this approach seems to lead to a higher score in every task. Therefore, the domain adaptation method inspired by FinBERT's training strategy, proves to be effective for decoder-only LLMs and not only for financial sentiment analysis, but for multiple financial NLP tasks.

\subsection{Comparison with other Financial LLMs}

\begin{table*}[ht!]
	\small
	\centering
	\begin{tabular}{p{2.2cm}cccc}
		\toprule
		             & \multicolumn{4}{c}{\textbf{F1 scores}}                                \\
		\cmidrule(lr){2-5}
		Models       & FPB                                    & FIQA-SA & Headlines & NER    \\
		\midrule
		BloombergGPT & 0.51                                   & 0.75    & 0.82      & 0.61   \\
		GPT-4        & 0.78                                   & -       & 0.86      & 0.83   \\
		FinMA-7B     & 0.86                                   & 0.84    & 0.98      & 0.75   \\
		FinMA-30B    & 0.88                                   & 0.87    & 0.97      & 0.62   \\
		\arrayrulecolor{black!30}\midrule
		Pythia-1.4B  & 0.84                                 & 0.83   & 0.97    & 0.69 \\
		OPT-1.3B     & 0.86                                 & 0.81  & 0.96     & 0.34  \\
		\arrayrulecolor{black!}\bottomrule
	\end{tabular}
	\caption{Comparison of state-of-the-art with Pythia-1.4B and OPT-1.3B fine-tuned on documents and the augmented dataset. Performance of models retrieved from BloombergGPT and PIXIU papers. BloombergGPT is not a publicly available model, so it is not possible to evaluate it under the same conditions as the other models. Thus, ChatGPT, GPT-4 and FinMA-30B are evaluated on zero-shot, BloombergGPT is only reported in a five-shot setting and its accuracy was not published.}\label{tab:sota-results}
\end{table*}

In this section, the results of the best models obtained through the previous experiments (\textit{Docs + Augmented instr}) are compared against the state-of-the-art LLMs for finance. As can be seen in Table~\ref{tab:sota-results}, both fine-tuned Pythia-1.4B and OPT-1.3B over perform GPT-4 in classification tasks, which includes financial sentiment analysis. This is made possible because of the domain adaptation conducted for these two base models. For NER, which is a generative task, GPT-4 is still the LLM with the highest score. When the models of these projects are compared to BloombergGPT, the biggest current LLM tailored for finance, it can be observed that the scores obtained are much higher for classification tasks, specially for sentiment analysis, and that Pythia also obtains better score for NER\@. In terms of efficiency, these results are achieved with models that have approximately $97\%$ fewer training parameters than BloombergGPT\footnote{BloombergGPT has 50B trainable parameters. Pythia-1.4B and OPT-1.3B have approximately 1.5B parameters. The amount of data used is not comparable since BloombergGPT was trained from scratch.}. In the comparative with the collection of FinMA models, the PIXIU LLMs still outperform the models fine-tuned with our domain adaptation strategy in some tasks, specially when compared to FinMA-30B. However, when FinMA-7B, the model with the closest size to the models presented in this project, is evaluated in financial sentiment analysis and Headlines, it can be observed that the scores are almost equivalent to the fine-tuned Pythia-1.4B and OPT-1.3B. In this case, however, the biggest improvement with respect to FinMA-7B is in terms of efficiency. Pythia-1.4B and OPT-1.3B have approximately $78\%$ fewer training parameters than FinMA-7B, and the number of instructions used goes from $220,226$ down to $126,909$, which is only a $57\%$ of the number of samples used for PIXIU models.

Therefore, from the general comparison it can be seen that the models fine-tuned in this project over perform most LLMs in financial tasks, with the only exception of FinMA models. In addition, the size of the models and the training strategy have been proven to be more efficient than the ones proposed for other models.

\section{Conclusion}\label{conclusion}
This project has covered a wide range of aspects of financial LLMs. Through a series of experiments, using Pythia-1.4B and OPT-1.3B as base models, we studied the adaptation of relatively small LLMs for finance. The experiments we conducted first show that LLMs adapted to the financial domain through further pre-training followed by instruction fine-tuning perform better than some of the best current generalist LLMs (such as GPT-4) on financial tasks. Second, it validates our training strategy since our LLMs obtain higher or similar scores than other financial LLMs that were trained with much more parameters and larger datasets.
%The experiments proved that LLMs adapted to this domain through further pre-training followed by instruction fine-tuning can lead them to perform better for financial tasks than some of the best current generalist LLMs, such as GPT-4, and to obtain higher or similar scores than other financial LLMs that were trained with more parameters and larger datasets.

Lowering the requirements to fine-tune LLMs for this specific industry can be key for the future of several companies, since it can enable smaller organizations to host their own LLMs or, at least, to make them more accessible. Furthermore, it is worth mentioning that the models used for this project as well as most datasets, except for the in-house subset (only $9.7\%$ of the documents dataset), are open and publicly available.
In addition to the findings related to domain adaptation of LLMs for financial tasks and the models presented, a strategy for the generation of samples for the instructions dataset is introduced. Moreover, the two datasets used for the project are described with enough details to be reproduced by other researchers. Finally, the paper also presented a comprehensive study that delves into the state-of-the-art and the evolution of approaches for financial sentiment analysis, ranging from traditional dictionary-based methods to the more recent advancements in LLMs.

Despite the fact that the results showed great performance of the small-sized models, in further research these fine-tuning strategies could be applied to larger models and study their impact on different scales and domains. An interesting option to study in the future are Low-Rank Adapters or LoRA~\cite{lora}, a method that reduces the number of trainable parameters by freezing the foundation model weights and injecting trainable rank decomposition matrices into each layer of the LLM\@.

\section*{Limitations}
The limitations of this work can be summarized as following:
\begin{itemize}
    \item Generative capabilities: The final fine-tuned model seems to perform very well on classification tasks such as sentiment analysis, while still lagging behind in generative ones.
    \item Unseen tasks: our work concentrates on certain tasks that have been studied in previous similar work, but for a full understanding of its limitations, one needs to test it on unseen tasks.
    \item Large models: we believe that testing the same strategy of multiple fine-tuning stages would yield even better results with larger models such as LLaMA-2-7B or even larger models.
\end{itemize}

\bibliography{main}

\begin{thebibliography}{21}
\expandafter\ifx\csname natexlab\endcsname\relax\def\natexlab#1{#1}\fi

\bibitem[{Araci(2019)}]{finbert}
Dogu Araci. 2019.
\newblock \href {http://arxiv.org/abs/arXiv:1908.10063} {Finbert: Financial
  sentiment analysis with pre-trained language models}.

\bibitem[{Biderman et~al.(2023)Biderman, Schoelkopf, Anthony, Bradley, O'Brien,
  Hallahan, Khan, Purohit, Prashanth, Raff, Skowron, Sutawika, and van~der
  Wal}]{pythia-llm}
Stella Biderman, Hailey Schoelkopf, Quentin Anthony, Herbie Bradley, Kyle
  O'Brien, Eric Hallahan, Mohammad~Aflah Khan, Shivanshu Purohit, USVSN~Sai
  Prashanth, Edward Raff, Aviya Skowron, Lintang Sutawika, and Oskar van~der
  Wal. 2023.
\newblock \href {http://arxiv.org/abs/2304.01373} {Pythia: A suite for
  analyzing large language models across training and scaling}.

\bibitem[{Devlin et~al.(2018)Devlin, Chang, Lee, and Toutanova}]{bert}
Jacob Devlin, Ming-Wei Chang, Kenton Lee, and Kristina Toutanova. 2018.
\newblock \href {http://arxiv.org/abs/arXiv:1810.04805} {Bert: Pre-training of
  deep bidirectional transformers for language understanding}.

\bibitem[{Gao et~al.(2020)Gao, Biderman, Black, Golding, Hoppe, Foster, Phang,
  He, Thite, Nabeshima, Presser, and Leahy}]{the-pile}
Leo Gao, Stella Biderman, Sid Black, Laurence Golding, Travis Hoppe, Charles
  Foster, Jason Phang, Horace He, Anish Thite, Noa Nabeshima, Shawn Presser,
  and Connor Leahy. 2020.
\newblock \href {http://arxiv.org/abs/2101.00027} {The pile: An 800gb dataset
  of diverse text for language modeling}.

\bibitem[{Howard and Ruder(2018)}]{ulmfit}
Jeremy Howard and Sebastian Ruder. 2018.
\newblock \href {http://arxiv.org/abs/1801.06146} {Universal language model
  fine-tuning for text classification}.

\bibitem[{Hu et~al.(2021)Hu, Shen, Wallis, Allen-Zhu, Li, Wang, Wang, and
  Chen}]{lora}
Edward~J. Hu, Yelong Shen, Phillip Wallis, Zeyuan Allen-Zhu, Yuanzhi Li, Shean
  Wang, Lu~Wang, and Weizhu Chen. 2021.
\newblock \href {http://arxiv.org/abs/2106.09685} {Lora: Low-rank adaptation of
  large language models}.

\bibitem[{Loshchilov and Hutter(2019)}]{adamw}
Ilya Loshchilov and Frank Hutter. 2019.
\newblock \href {http://arxiv.org/abs/1711.05101} {Decoupled weight decay
  regularization}.

\bibitem[{Maia et~al.(2018)Maia, Handschuh, Freitas, Davis, McDermott, Zarrouk,
  and Balahur}]{fiqasa}
Macedo Maia, Siegfried Handschuh, Andr{\'e} Freitas, Brian Davis, Ross
  McDermott, Manel Zarrouk, and Alexandra Balahur. 2018.
\newblock Www'18 open challenge: financial opinion mining and question
  answering.
\newblock In \emph{Companion proceedings of the the web conference 2018}, pages
  1941--1942.

\bibitem[{Malo et~al.(2014)Malo, Sinha, Korhonen, Wallenius, and
  Takala}]{financial-phrasebank}
Pekka Malo, Ankur Sinha, Pekka Korhonen, Jyrki Wallenius, and Pyry Takala.
  2014.
\newblock Good debt or bad debt: Detecting semantic orientations in economic
  texts.
\newblock \emph{Journal of the Association for Information Science and
  Technology}, 65(4):782--796.

\bibitem[{OpenAI(2023)}]{gpt4}
OpenAI. 2023.
\newblock \href {http://arxiv.org/abs/2303.08774} {Gpt-4 technical report}.

\bibitem[{Radford et~al.(2019)Radford, Wu, Child, Luan, Amodei, Sutskever
  et~al.}]{in-context-learning}
Alec Radford, Jeffrey Wu, Rewon Child, David Luan, Dario Amodei, Ilya
  Sutskever, et~al. 2019.
\newblock Language models are unsupervised multitask learners.
\newblock \emph{OpenAI blog}, 1(8):9.

\bibitem[{Salinas~Alvarado et~al.(2015)Salinas~Alvarado, Verspoor, and
  Baldwin}]{ner}
Julio~Cesar Salinas~Alvarado, Karin Verspoor, and Timothy Baldwin. 2015.
\newblock Domain adaption of named entity recognition to support credit risk
  assessment.
\newblock In \emph{Proceedings of the Australasian Language Technology
  Association Workshop 2015}, pages 84--90, Parramatta, Australia.

\bibitem[{Scao et~al.(2023)Scao, Fan, Akiki, Pavlick, Ilić, Hesslow,
  Castagné, Luccioni, Yvon, Gallé, and et~al}]{bloom-short}
Teven~Le Scao, Angela Fan, Christopher Akiki, Ellie Pavlick, Suzana Ilić,
  Daniel Hesslow, Roman Castagné, Alexandra~Sasha Luccioni, François Yvon,
  Matthias Gallé, and et~al. 2023.
\newblock \href {http://arxiv.org/abs/2211.05100} {Bloom: A 176b-parameter
  open-access multilingual language model}.

\bibitem[{Sinha and Khandait(2020)}]{gold}
Ankur Sinha and Tanmay Khandait. 2020.
\newblock \href {http://arxiv.org/abs/2009.04202} {Impact of news on the
  commodity market: Dataset and results}.

\bibitem[{Touvron et~al.(2023{\natexlab{a}})Touvron, Lavril, Izacard, Martinet,
  Lachaux, Lacroix, Rozière, Goyal, Hambro, Azhar, Rodriguez, Joulin, Grave,
  and Lample}]{llama}
Hugo Touvron, Thibaut Lavril, Gautier Izacard, Xavier Martinet, Marie-Anne
  Lachaux, Timothée Lacroix, Baptiste Rozière, Naman Goyal, Eric Hambro,
  Faisal Azhar, Aurelien Rodriguez, Armand Joulin, Edouard Grave, and Guillaume
  Lample. 2023{\natexlab{a}}.
\newblock \href {http://arxiv.org/abs/arXiv:2302.13971} {Llama: Open and
  efficient foundation language models}.

\bibitem[{Touvron et~al.(2023{\natexlab{b}})Touvron, Martin, Stone, Albert,
  Almahairi, Babaei, Bashlykov, Batra, Bhargava, Bhosale, Bikel, Blecher,
  Ferrer, Chen, Cucurull, Esiobu, Fernandes, Fu, Fu, Fuller, Gao, Goswami,
  Goyal, Hartshorn, Hosseini, Hou, Inan, Kardas, Kerkez, Khabsa, Kloumann,
  Korenev, Koura, Lachaux, Lavril, Lee, Liskovich, Lu, Mao, Martinet, Mihaylov,
  Mishra, Molybog, Nie, Poulton, Reizenstein, Rungta, Saladi, Schelten, Silva,
  Smith, Subramanian, Tan, Tang, Taylor, Williams, Kuan, Xu, Yan, Zarov, Zhang,
  Fan, Kambadur, Narang, Rodriguez, Stojnic, Edunov, and Scialom}]{llama2-llm}
Hugo Touvron, Louis Martin, Kevin Stone, Peter Albert, Amjad Almahairi, Yasmine
  Babaei, Nikolay Bashlykov, Soumya Batra, Prajjwal Bhargava, Shruti Bhosale,
  Dan Bikel, Lukas Blecher, Cristian~Canton Ferrer, Moya Chen, Guillem
  Cucurull, David Esiobu, Jude Fernandes, Jeremy Fu, Wenyin Fu, Brian Fuller,
  Cynthia Gao, Vedanuj Goswami, Naman Goyal, Anthony Hartshorn, Saghar
  Hosseini, Rui Hou, Hakan Inan, Marcin Kardas, Viktor Kerkez, Madian Khabsa,
  Isabel Kloumann, Artem Korenev, Punit~Singh Koura, Marie-Anne Lachaux,
  Thibaut Lavril, Jenya Lee, Diana Liskovich, Yinghai Lu, Yuning Mao, Xavier
  Martinet, Todor Mihaylov, Pushkar Mishra, Igor Molybog, Yixin Nie, Andrew
  Poulton, Jeremy Reizenstein, Rashi Rungta, Kalyan Saladi, Alan Schelten, Ruan
  Silva, Eric~Michael Smith, Ranjan Subramanian, Xiaoqing~Ellen Tan, Binh Tang,
  Ross Taylor, Adina Williams, Jian~Xiang Kuan, Puxin Xu, Zheng Yan, Iliyan
  Zarov, Yuchen Zhang, Angela Fan, Melanie Kambadur, Sharan Narang, Aurelien
  Rodriguez, Robert Stojnic, Sergey Edunov, and Thomas Scialom.
  2023{\natexlab{b}}.
\newblock \href {http://arxiv.org/abs/2307.09288} {Llama 2: Open foundation and
  fine-tuned chat models}.

\bibitem[{Vaswani et~al.(2023)Vaswani, Shazeer, Parmar, Uszkoreit, Jones,
  Gomez, Kaiser, and Polosukhin}]{transformers}
Ashish Vaswani, Noam Shazeer, Niki Parmar, Jakob Uszkoreit, Llion Jones,
  Aidan~N. Gomez, Lukasz Kaiser, and Illia Polosukhin. 2023.
\newblock \href {http://arxiv.org/abs/1706.03762} {Attention is all you need}.

\bibitem[{Wu et~al.(2023)Wu, Irsoy, Lu, Dabravolski, Dredze, Gehrmann,
  Kambadur, Rosenberg, and Mann}]{bloomberggpt}
Shijie Wu, Ozan Irsoy, Steven Lu, Vadim Dabravolski, Mark Dredze, Sebastian
  Gehrmann, Prabhanjan Kambadur, David Rosenberg, and Gideon Mann. 2023.
\newblock \href {http://arxiv.org/abs/arXiv:2303.17564} {Bloomberggpt: A large
  language model for finance}.

\bibitem[{Xie et~al.(2023)Xie, Han, Zhang, Lai, Peng, Lopez-Lira, and
  Huang}]{finma}
Qianqian Xie, Weiguang Han, Xiao Zhang, Yanzhao Lai, Min Peng, Alejandro
  Lopez-Lira, and Jimin Huang. 2023.
\newblock \href {http://arxiv.org/abs/arXiv:2306.05443} {Pixiu: A large
  language model, instruction data and evaluation benchmark for finance}.

\bibitem[{Zhang et~al.(2022)Zhang, Roller, Goyal, Artetxe, Chen, Chen, Dewan,
  Diab, Li, Lin, Mihaylov, Ott, Shleifer, Shuster, Simig, Koura, Sridhar, Wang,
  and Zettlemoyer}]{opt-llm}
Susan Zhang, Stephen Roller, Naman Goyal, Mikel Artetxe, Moya Chen, Shuohui
  Chen, Christopher Dewan, Mona Diab, Xian Li, Xi~Victoria Lin, Todor Mihaylov,
  Myle Ott, Sam Shleifer, Kurt Shuster, Daniel Simig, Punit~Singh Koura, Anjali
  Sridhar, Tianlu Wang, and Luke Zettlemoyer. 2022.
\newblock \href {http://arxiv.org/abs/2205.01068} {Opt: Open pre-trained
  transformer language models}.

\bibitem[{Zhou et~al.(2023)Zhou, Liu, Xu, Iyer, Sun, Mao, Ma, Efrat, Yu, Yu,
  Zhang, Ghosh, Lewis, Zettlemoyer, and Levy}]{lima}
Chunting Zhou, Pengfei Liu, Puxin Xu, Srini Iyer, Jiao Sun, Yuning Mao, Xuezhe
  Ma, Avia Efrat, Ping Yu, Lili Yu, Susan Zhang, Gargi Ghosh, Mike Lewis, Luke
  Zettlemoyer, and Omer Levy. 2023.
\newblock \href {http://arxiv.org/abs/2305.11206} {Lima: Less is more for
  alignment}.

\end{thebibliography}
\bibliographystyle{acl_natbib}

\appendix

\section{Instruction data augmentation examples}\label{sec:inst_data_aug}

\begin{table}[ht]
	\small
	\centering
	\begin{tabular}{p{0.94\linewidth}}
		\toprule
		\textbf{Template}                                                                                                                                                                                                                                                                                                                                            \\
		\midrule
		\texttt{Write a sentence with a $\{ y_i \}$ financial sentiment. Use the format <stc> sentence </stc>. Reuse terms from the example. Example: '<stc> $\{ x_i \}$ </stc>'}                                                                                                                                                                                    \\
		\midrule
		\textbf{Example}                                                                                                                                                                                                                                                                                                                                             \\
		\midrule
		\texttt{Write a sentence with a \textbf{positive} financial sentiment. Use the format <stc> sentence </stc>. Reuse terms from the example. Example: '<stc> \textbf{Shares of Standard Chartered ( STAN ) rose 1.2 \% in the FTSE 100 , while Royal Bank of Scotland ( RBS ) shares rose 2 \% and Barclays shares ( BARC ) ( BCS ) were up 1.7 \% .} </stc>'} \\
		\bottomrule
	\end{tabular}
	\caption{Template for sentiment analysis input generation with financial sentiment fixed and dynamic shot. Example given for positive input inference. $\{ xi, yi \}$ is an \textit{input-answer} pair sampled from one of the two subsets.}\label{tab:template-sa}
\end{table}

\begin{table}[ht]
	\small
	\centering
	\begin{tabular}{p{0.94\linewidth}}
		\toprule
		\textbf{Template}                                                                                                                                                                                                                                                                                                                        \\
		\midrule
		\texttt{Identify the named entities that represent a person ('PER'), an organization ('ORG'), or a location ('LOC') in a financial context. Use the format 'Entities: entity name, entity type'.}

		\texttt{Sentence: 'The Bank gave money to the Borrower to open a business in New York.'; Entities: 'Bank, ORG | Borrower, PER | New York, LOC'}

		\texttt{Do the same with this sentence, identifying 'PER', 'ORG', 'LOC' entities.}

		\texttt{Sentence: $\{x_i\}$; Entities:}                                                                                                                                                                                                                                                                                                  \\
		\midrule
		\textbf{Example}                                                                                                                                                                                                                                                                                                                         \\
		\midrule
		\texttt{Identify the named entities that represent a person ('PER'), an organization ('ORG'), or a location ('LOC') in a financial context. Use the format 'Entities: entity name, entity type'.}

		\texttt{Sentence: 'The Bank gave money to the Borrower to open a business in New York.'; Entities: 'Bank, ORG | Borrower, PER | New York, LOC'}

		\texttt{Do the same with this sentence, identifying 'PER', 'ORG', 'LOC' entities.}

		\texttt{Sentence: \textbf{‘350 , Wellesley , Massachusetts 02481 doing business as " Silicon Valley East " and AKAMAI TECHNOLOGIES , INC . (" Borrower "), whose address is 201 Broadway , 4th Floor , Cambridge , Massachusetts 02139 provides the terms on which Bank will lend to Borrower and Borrower will repay Bank’}; Entities:} \\
		\bottomrule
	\end{tabular}
	\caption{Template for NER tags generation given a sentence of the financial domain.}\label{tab:template-ner}
\end{table}

\end{document}